\pdfoutput=1

\documentclass[11pt]{article}

\usepackage[final]{acl}

\usepackage{times}
\usepackage{latexsym}

\usepackage[T1]{fontenc}

\usepackage[utf8]{inputenc}

\usepackage{microtype}

\usepackage{inconsolata}

\usepackage{graphicx}

\makeatletter
\DeclareRobustCommand\onedot{\futurelet\@let@token\@onedot}
\def\@onedot{\ifx\@let@token.\else.\null\fi\xspace}

\newcommand{\eg}{\textit{e.g.}}
\newcommand{\ie}{\textit{i.e.}}

\makeatother
\usepackage{colortbl}
\usepackage{amsmath}
\usepackage{amssymb}
\usepackage{mathtools}
\usepackage{adjustbox}
\usepackage{marvosym}
\usepackage{colortbl}
\usepackage{multirow}
\usepackage{graphicx}
\usepackage{booktabs}
\usepackage{caption}
\usepackage{subcaption}
\usepackage{tabularx}
\usepackage{pifont}
\usepackage{arydshln}
\usepackage{makecell}

\usepackage{xcolor}

\definecolor{highlightcolor}{gray}{.9}

\definecolor{baselinecolor}{gray}{.9}
\definecolor{backcolor}{RGB}{232, 242, 255}

\title{Mixture of Cluster-Conditional LoRA Experts \\
for Vision-Language Instruction Tuning}

\author{
 \textbf{Yunhao Gou\textsuperscript{1,2*}},
 \textbf{Zhili Liu\textsuperscript{2,3*}},
 \textbf{Kai Chen\textsuperscript{2*}},
 \textbf{Lanqing Hong\textsuperscript{3}},
 \textbf{Hang Xu\textsuperscript{3}},
 \textbf{Aoxue Li\textsuperscript{3}},
\\
 \textbf{Xin Jiang\textsuperscript{3}},
 \textbf{Zhenguo Li \textsuperscript{3}},
 \textbf{Dit-Yan Yeung\textsuperscript{2}},
 \textbf{James T. Kwok\textsuperscript{2}},
 \textbf{Yu Zhang\textsuperscript{1,4}},
\\
\\
 \textsuperscript{1}Southern University of Science and Technology,
 \\
 \textsuperscript{2}The Hong Kong University of Science and Technology,
 \\
 \textsuperscript{3}Huawei Noah’s Ark Lab,
 \textsuperscript{4}Peng Cheng Laboratory,
\\
 \small{
   \textbf{$^*$} Equal contribution.
   \quad
   \textbf{Correspondence:} \href{mailto:yu.zhang.ust@gmail.com}{yu.zhang.ust@gmail.com}
 }
}

\begin{document}
\maketitle
\begin{abstract}
Instruction tuning of Large Vision-language Models (LVLMs) has revolutionized the development of versatile models with zero-shot generalization across a wide range of downstream vision-language tasks.
However, the diversity of training tasks of different sources and formats would lead to inevitable task conflicts, where different tasks conflict for the same set of model parameters, resulting in sub-optimal instruction-following abilities.
To address that, we propose the Mixture of Cluster-conditional LoRA Experts (MoCLE), a novel Mixture of Experts (MoE) architecture designed to activate the task-customized model parameters based on the instruction clusters. 
A separate universal expert is further incorporated to improve generalization capabilities of MoCLE for novel instructions.
Extensive experiments on InstructBLIP and LLaVA demonstrate the effectiveness of MoCLE.
\end{abstract}


\section{Introduction}
\label{sec_intro}
There has been a continuously increasing trend to develop intelligent assistants that can follow human instructions~\cite{brown2020language,chatgpt,chen2023gaining}, 
with instruction tuning emerging as a notably effective approach. 
This method leverages large-scale well-formatted instruction data to empower Large Language Models (LLMs) to execute various human instructions, showcasing their ability to generalize across novel unseen tasks~\cite{longpre2023flan}. 
Likewise, efforts have been made to introduce similar capabilities to Large Vision-language Models (LVLMs) \cite{Bai2023QwenVLAV, zhang2023internlm, ye2023mplug, chen2023sharegpt4v, chen2023minigpt}, including LLaVA series \cite{Liu2023VisualIT, liu2023improved}, MiniGPT-4~\cite{Zhu2023MiniGPT4EV} and InstructBLIP~\cite{Dai2023InstructBLIPTG}. 

It is observed that for both the LLMs~\cite{sanh2021multitask,wang2022super,chung2022scaling} and LVLMs~\cite{Bai2023QwenVLAV,Zhao2023SVITSU,Li2023M3ITAL}, the ability to generalize to novel unseen instructions necessitates multi-task instruction tuning, \ie, training on a diverse collection of instruction-following tasks.
However, the complexity of various instruction tasks brings difficulties for model fine-tuning. 
Specifically, 
\citeauthor{wei2021finetuned} 
(\citeyear{wei2021finetuned})
find that for certain model sizes, multi-task instruction tuning even fails to bring performance gains for zero-shot tasks compared to the original models. This is mainly attributed to the \textit{negative transfer} phenomenon~\cite{Zhang2017ASO} during multi-task instruction tuning, where the model struggles to optimize the losses of multiple conflicted tasks, leading to sub-optimal performance. 


\begin{figure}[t]
\centering
\vspace{-1em}
\includegraphics[width=\columnwidth]{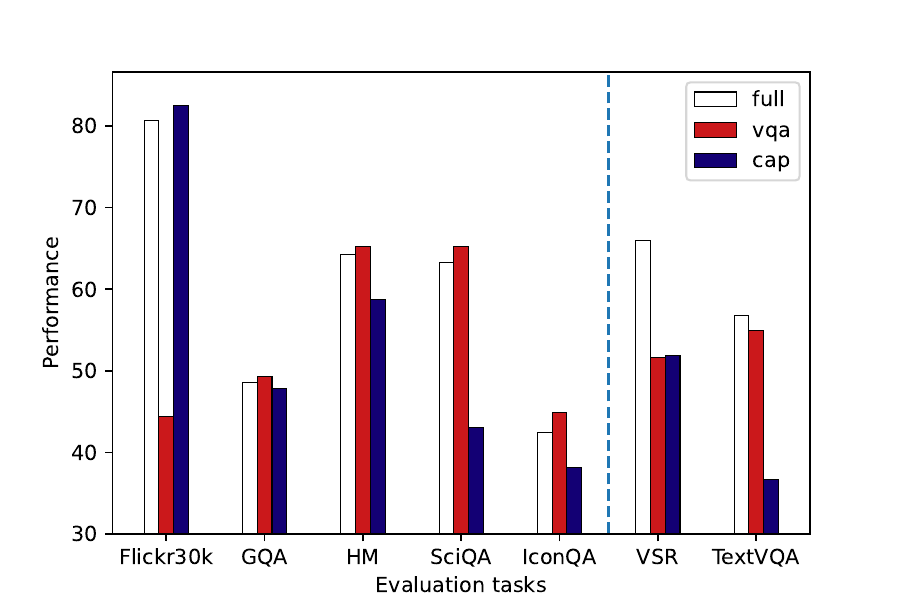}
\vspace{-1em}
\caption{\textbf{Performance of the instruction-finetuned LVLMs on zero-shot tasks}, where larger values indicate better performance. Only 2 out of 7 tasks benefit from instruction tuning from all the data, while the task experts show better performance on the other 5 tasks (\ie, Flickr 30K, GQA, HM, SciQA and IconQA).}
\label{fig_pre}
\end{figure}


Similarly, tasks for vision-language instruction tuning (\eg, visual question answering and image captioning) focus on different perspectives of LVLMs. 
This results in conflicts
as most studies adopt
sharing of all parameters.
In our preliminary study, we split the instruction data into two \textit{disjoint} subsets (\emph{``cap''} for image captioning, and \emph{``vqa''} for visual question answering). 
We then train InstructBLIP~\cite{Dai2023InstructBLIPTG}  using LoRA~\cite{Hu2021LoRALA} on three data sets (\emph{``cap''}, \emph{``vqa''} and 
the full data
\emph{``full''}) to obtain three sets of parameters (\ie, task experts). 
Following the held-out evaluation protocol~\cite{Dai2023InstructBLIPTG}, we evaluate these experts on the unseen datasets/tasks with the best expert.  
As shown in Figure \ref{fig_pre}, on 5 out of the 7 downstream tasks, the InstructBLIP instruction-tuned on all the data is outperformed by the task expert finetuned with only a subset of data.
Among the 5 tasks, Flickr30k belongs to \emph{``cap''}, and SciQA, GQA and IconQA belong to \emph{``vqa''}. This shows that \textit{instruction tuning on similar tasks brings positive transfer to downstream tasks, while training on the full data with dissimilar tasks can hurt generalization performance}.

The use of disjoint task experts above is a na\"ive solution to negative transfer, where we manually partition the training tasks and train each expert separately. However, it has several limitations:
(1) The taxonomies such as \emph{``vqa''} and \emph{``cap''} require human expertise, and are difficult to scale as the number of tasks grows.
(2) The ability to generalize to unseen tasks is inhibited, as we do not know which expert to choose for novel tasks, while some new tasks might benefit from multiple training tasks (\eg, VSR and TextVQA as in Figure~\ref{fig_pre}). 
In this regard, specialization and generalization of LVLMs becomes a dilemma. 

This paper aims to develop an automatic and practical partition strategy and a network architecture that strikes a balance between specialization and generalization.
In particular, we propose the \textit{Mixture of Cluster-conditional LoRA Experts} (MoCLE) for vision-language instruction tuning. 
In this proposed framework, we first cluster instructions of all the training data into several clusters via a pre-trained clustering model. In this way, similar tasks that can bring positive transfer to each other are automatically grouped into the same cluster, while different tasks that may cause conflict are separated (more justifications for the use of instruction clusters are detailed in Sec.~\ref{sec_cluster}). 
Then we construct several task experts, with each focusing on a specific cluster. Using the cluster as condition,  
a router dispatches the input data to one of the specialized task experts and an universal expert that is shared among all data. 
As we activate a specialized expert for a group of similar tasks, tasks that are less similar are learned via separate experts, mitigating task conflicts. 
Meanwhile, since the universal expert trained on all tasks also contributes to the model outputs, we can enjoy generalization and specialization simultaneously. 

We validate effectiveness of MoCLE on InstructBLIP~\cite{Dai2023InstructBLIPTG} and LLaVA-1.5 \cite{liu2023improved} and observe remarkable performance gains compared to dense models and other MoE baselines\cite{chen2024octavius,chen2024llava}. 

The main contributions of this work contain the following three parts,
\begin{enumerate}
    \item We identify the negative transfer phenomenon \cite{liu2022task,zhili2023task} as tasks conflict during instruction tuning of LVLMs.
    \item We propose \textit{Mixture of Cluster-conditional LoRA Experts} (MoCLE), a novel parameter-efficient finetuning framework suitable for the vision-language instruction tuning, to mitigate task conflicts and enjoy the benefits of huge data training simultaneously.
    \item Our proposed MoCLE achieves remarkable performance gains on held-in/out tasks compared to dense models and other MoE baselines\cite{chen2024octavius,chen2024llava}.
\end{enumerate}


\section{Related Work}

\subsection{Multi-Task Instruction Tuning}
Instruction tuning \cite{sanh2021multitask, wei2021finetuned} fine-tunes a language model across tasks and instruction templates to convey task intentions. Its goal is to teach the model to understand relationships between instructions and input/output pairs, enabling generalization to unseen tasks with novel instructions. Increasing the number of instructions \cite{sanh2021multitask}, tasks \cite{wang2022super, chung2022scaling}, and data diversity \cite{zhou2023lima} have shown to be effective in improving performance. However, \citeauthor{wei2021finetuned} (\citeyear{wei2021finetuned}) find that for certain model sizes, instruction tuning fails to outperform untuned models on unseen tasks due to full capacity utilization for learning task mixtures. Our work addresses this in vision-language instruction tuning using specialized experts.

\begin{figure*}[!t]
\centering
\includegraphics[width=\linewidth]{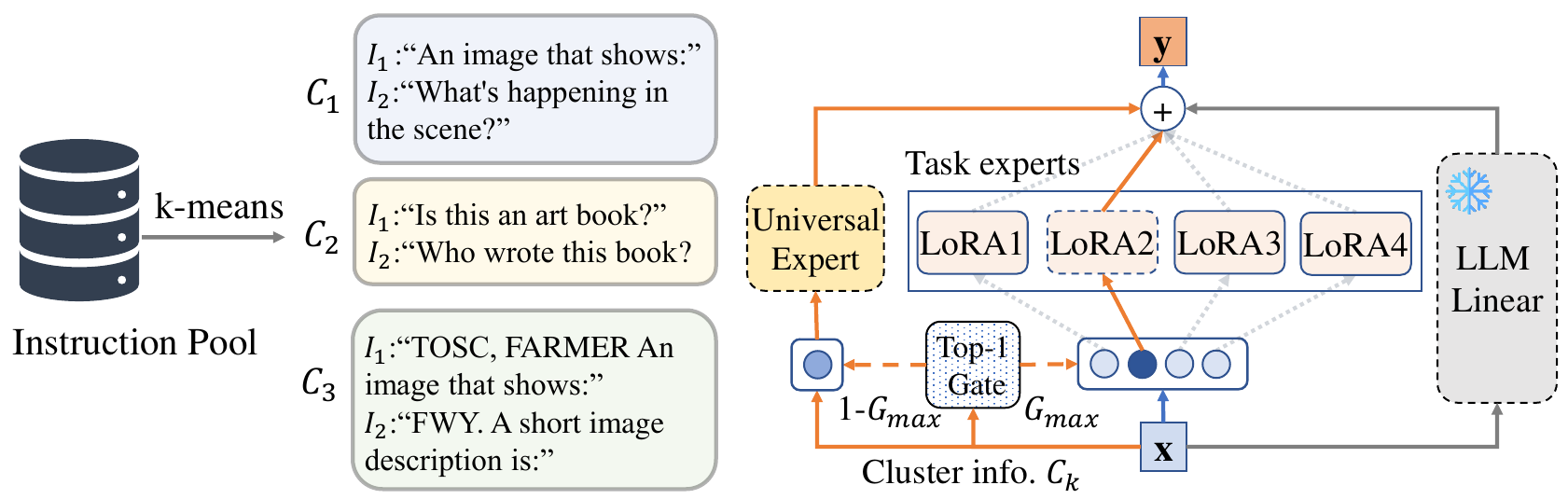}
\vspace{-3mm}
\caption{\textbf{Overall pipeline of MoCLE.}}
\label{fig_method}
\end{figure*}


\subsection{Mixture of Experts (MoE)}
MoE models \cite{Jacobs1991AdaptiveMO,Jordan1993HierarchicalMO,Shazeer2017OutrageouslyLN} are renowned for their ability to increase model capacity through parameter expansion. Recent research integrates MoE with adapters, exploring how pretrained adapters can be effectively combined \cite{wu2024mixture}, and how they enhance performance in both few-shot \cite{huang2023lorahub} and zero-shot scenarios \cite{Jang2023ExploringTB,muqeeth2024learning}.
Another line of research \cite{chen2024llava,wu2024parameter,luo2024moelora,Zadouri2023PushingMO,chen2024octavius} focuses on augmenting model capacity in a parameter-efficient manner. However, these approaches do not explicitly incorporate task/domain priors during the routing process, which might be limited when handling task conflicts.
Two concurrent studies \cite{liu2024intuition, li2024mixlora} incorporate domain information for expert routing. Unlike our approach, however, they do not address zero-shot generalization to unseen tasks.


\section{Methodology}

In this section, we start with the formulation of LVLM instruction tuning and an analysis of the limitations of task experts. We then introduce the proposed MoCLE. The overall framework is shown in Figure \ref{fig_method}.


\subsection{Problem Formulation}
\label{sec_problem}
Suppose that there is a set of datasets that are divided into held-in and held-out datasets~\cite{Dai2023InstructBLIPTG}. A large vision-language model is first finetuned on the held-in dataset, and then evaluated on the held-out dataset in a zero-shot manner. 
To unify and diversify input-output formats and promotes instruction tuning, several task templates $\{T_{i}\}$ are designed to wrap the raw inputs, which is a pair of text 
$X_\text{txt}$
and image 
$X_\text{img}$ from the dataset. 
For example, ``\textit{Given the image, answer the question with no more than three words.} \{Question\}'' is a template for visual question answering tasks. 
The instruction is defined as $I \equiv T_{i}(X_\text{txt})$ that wraps text inputs using the template. 


\subsection{Clustering Data by Instructions}
\label{sec_cluster}

The purposes of partitioning the training data are two-fold. 
First, we hope to train a task expert with a collection of similar tasks so as to avoid task conflicts. 
Second, we expect novel tasks to be automatically assigned to the proper experts based on their cluster without manual intervention.

To achieve these goals, we conduct clustering on the instructions as they serve as the foundation for identifying different tasks. 
Formally, let $\mathcal{E}(\cdot)$ be a pre-trained sentence encoder, and $\mathbf{e}_{i} = \mathcal{E}(I_i)$ be the sentence representation
of an instruction $I_i$. We use the $k$-means clustering algorithm to group all instructions in the training datasets into $K$ clusters by iteratively minimizing \( \sum_{j=1}^{K} \sum_{\mathbf{e}_i \in S_j} \|\mathbf{e}_i - \mathbf{c}_j\|^2 \), where \( S_j \) is the set of instructions assigned to the $j$th cluster, and $\mathbf{c}_j$ is the centroid of the $j$th cluster. In each 
$k$-means clustering 
iteration, each instruction is assigned to the nearest centroid with all\ centroids updated as the average of instruction representations in the corresponding cluster.


\subsection{Mixture of Cluster-Conditional LoRA Experts}
\label{sec_moe}
In addition to considerations at the data level, we also suggest an architectural design to tackle the issue of negative transfer.
We propose the Mixture of Cluster-conditional LoRA Experts (MoCLE) that learns to activate the LoRA expert at each layer given the cluster of the data.
Specifically, let $E$ as be the number of experts. We introduce a gate vector $\mathbf{G} \in \mathbb{R}^E$. Given an input $\mathbf{x}_i$, 
$\mathbf{G}$
determines the experts to which the input is routed. The gate vector is obtained as:
\begin{equation}
\label{eq_moe_gate}
    \mathbf{G} = \text{top}_k \left( \text{softmax}\left(\frac{1}{\tau}\left(\mathbf{W}_{\text{gate}} \mathbf{C}_{[\mathbf{x}_i]} + \boldsymbol{\epsilon}\right)\right) \right),
\end{equation}
where $\text{top}_k(\cdot)$ keeps the $k$ largest entries unchanged and sets the others to zero. $\mathbf{C}_{[\mathbf{x}]}$, which is shared among all layers, is the learnable embedding of the cluster that $\mathbf{x}$ belongs to. 
This is the key for the model to choose proper task experts for the input data. 
To endow the clustering embedding with task information, we initialize it to be the centroid of the corresponding cluster.
Moreover, $\mathbf{W}_{\text{gate}}$ is the trainable weights of the linear gate, which is learned at each layer where the MoE block is inserted, $\boldsymbol{\epsilon} \sim N(0, \frac{1}{E})$ is a noise term that adds randomness to the expert choosing process (and encourages MoCLE to explore multiple combinations of experts
during training\footnote{We do not apply load balancing during training as we found it might distort task specialization.}), and $\tau$ is a temperature hyperparameter.
The output $\mathbf{y}_i$ is then computed as the sum of weighted outputs of the experts, and the original LLM linear layer~\cite{Hu2021LoRALA} on the input $\mathbf{x}_i$, as:
\begin{equation}
\label{eq_moe_sum}
    \mathbf{y}_i = \sum_{e=1}^{E} G_e \mathbf{W}_e \mathbf{x}_i + \mathbf{W}_0\mathbf{x}_i,
\end{equation}
where $\mathbf{W}_0$ is the pre-trained linear layer of LVLM, $\mathbf{W}_e$ is the linear projection weight of the $e$th LoRA expert, and $G_e$ (the $e$th entry in $\mathbf{G}$) indicates the contribution of the $e$th expert. 


\begin{table*}[!t]
\begin{adjustbox}{width=\linewidth, center}
\begin{tabular}{@{}l|ccccccc@{}}
\toprule
Models & LLM & Expert Params. & \# Experts & \# Clusters & Rank & Temperature & Trainable Params. \\   \midrule
InstructBLIP & Vicuna-7B & q\_proj, v\_proj & 4 + 1(universal) & 64 & 8 & 0.05 & Q-Former, LoRAs \\
LLaVA-1.5 & Vicuna-1.5-7B & up\_proj, down\_proj & 4 + 1(universal) & 4 & 128 & 0.1 & MLP connector, LoRAs \\
\bottomrule
\end{tabular}%
\end{adjustbox}
\caption{Architecture details of MoCLE on different LVLMs. Note that experts are added to each layer of the transformer. }
\label{tbl_models}
\vspace{-4mm}
\end{table*}

\subsection{Universal Expert}
\label{sec_general}

As will be shown in Sec. \ref{sec_effect}, the formulation in Sec. \ref{sec_moe}
still hurts the generalization ability of the entire model, due to the fact that instruction-tuned models generalize to unseen tasks via training on extensive instructions~\cite{wei2021finetuned}, while in our formulation, each expert sees fewer instructions than the original dense model.

To alleviate this problem, we propose an \textit{universal expert} that learns from all training data. 
Specifically, we fix the number of activated experts to 1 (\ie, $k$ in Eq. \ref{eq_moe_gate} equals 1) and define $G_\text{max}$ as the maximum element in $\mathbf{G}$. Then the output for all the experts is expressed as:
\begin{equation} \label{eq:univ}
\mathbf{y}_i = \left(\sum_{e=1}^{E} G_e \mathbf{W}_e  + (1-G_{\text{max}}) \mathbf{W}_u \right) \mathbf{x}_i + \mathbf{W}_0\mathbf{x}_i.
\end{equation}
in which we additionally train an universal expert parameterized by $\mathbf{W}_u$. Different from the task experts that are activated only for specific model inputs, the universal expert is activated for all inputs. The final output is a weighted sum of outputs from one of the experts and the universal expert plus the original LVLM's output. 
Consequently, the task expert learns distinct skills for certain tasks while the universal expert masters holistic understanding of the training corpus. The synergy between them offers both specialization and generalization for the LVLMs with MoCLE.


\begin{table*}[!t]
\begin{adjustbox}{width=0.8\linewidth, center}
\begin{tabular}{@{}l|cccccccccc@{}}
\toprule
 \multirow{2}{*}{Models} &  \multirow{2}{*}{GQA} &  \multirow{2}{*}{VSR} &  \multirow{2}{*}{IQA} &  \multirow{2}{*}{Visdial} & \multirow{2}{*}{MME}  & \multirow{2}{*}{POPE} & \multicolumn{2}{c}{A-OKVQA} & OKVQA & VQAv2 \\ 
 & & & & & & & Direct & MC & (test) & (test-dev) \\\midrule 
 InstructBLIP (7B) & 48.6 & 60.8 & 43.4 & 46.3 & 1202.9 & 77.6 & 58.8 & 73.8 & 57.0 & 77.4 \\ 
 \rowcolor{backcolor} 
  + MoCLE & \textbf{49.3} & \textbf{64.7} & \textbf{46.3} & \textbf{46.9} & \textbf{1222.6} & \textbf{82.1} & \textbf{61.5} & \textbf{78.2} & \textbf{59.8} & \textbf{78.9} \\

\bottomrule
\end{tabular}%
\end{adjustbox}
\caption{\textbf{Zero-shot results (InstructBLIP) on the held-out datasets}, \ie, GQA, VSR, IconQA (IQA), Visdial, MME, POPE and \textbf{evaluation on held-in datasets}, \ie, A-OKVQA, OKVQA, VQAv2. Here Direct and MC denote directly answering and multiple choices. Best results are marked in \textbf{bold}. }
\label{tbl_main}
\vspace{-4mm}
\end{table*}


\begin{table*}[!t]
\begin{adjustbox}{width=0.7\linewidth, center}
\begin{tabular}{@{}l|ccccccc@{}}
\toprule
 Models &
  \setlength\extrarowheight{0pt}\begin{tabular}[c]{@{}c@{}}Flickr\\ 30K\end{tabular} &
   VQA$^{T}$ & 
   HM &
  SQA&
   \begin{tabular}[c]{@{}c@{}}MSVD\\ QA\end{tabular} &
   \begin{tabular}[c]{@{}c@{}}MSRVTT\\ QA\end{tabular} &
   iVQA \\ \midrule 

InstructBLIP (7B) & 81.3 
& {53.9} & {65.3} & {62.0} & 41.4 & {23.0} & 51.3 \\
\rowcolor{backcolor}
 + MoCLE  & \textbf{81.9} & \textbf{57.1} & \textbf{65.6} & \textbf{63.9} & \textbf{42.6} & 
\textbf{24.4} & 
\textbf{53.2} \\

\bottomrule
\end{tabular}%
\end{adjustbox}
\caption{
\textbf{Zero-shot results (InstructBLIP) on the held-out datasets.} 
Here, VQA$^{T}$, HM and SQA denote TextVQA, HatefulMemes and ScienceQA, respectively.
}
\label{tbl_main2}
\vspace{-4mm}
\end{table*}


\begin{table*}[!t]
\begin{adjustbox}{width=0.9\linewidth, center}
\begin{tabular}{@{}lc|ccc|c|cccccc@{}}
\toprule
 \multirow{2}{*}{Methods} & \multirow{2}{*}{Train Data} & \multirow{2}{*}{MME} & \multirow{2}{*}{MMB} & \multirow{2}{*}{SQA} & \multirow{2}{*}{GeoQA} & \multicolumn{2}{c}{VQA-RAD} & \multicolumn{2}{c}{SLAKE} & \multicolumn{2}{c}{PathVQA}  \\ 
 & & & & & & Open & Closed & Open & Closed & Open & Closed \\ \midrule 
 \multirow{3}{*}{Single LoRA} & LLaVA-665k & 1804 & 65.89 & 67.67 & - & - & - & - & - & - & - \\
  & Geo170k & - & - & - & 57.82 & - & - & - & - & - & -\\
  & Med. Mix & - & - & - & - & 53.90 & 84.19 & 86.05 & 85.58 & 38.07
& 91.77 
 \\\midrule
 Single LoRA & All & 1794 & 64.69 & 66.78 & 57.56 & 46.89 & 77.94 & \textbf{84.61} & 82.45 & \textbf{35.56} & 90.71 \\
 \rowcolor{backcolor} 
 MoCLE  & All & \textbf{1838} & \textbf{66.07} & \textbf{67.38} & \textbf{60.21} & \textbf{53.59} & \textbf{81.98} & 83.29 & \textbf{85.10} & 35.21 & \textbf{91.65}  \\
 
\bottomrule
\end{tabular}%
\end{adjustbox}
\caption{
\textbf{Evaluation results of LLaVA-1.5-7B}, where MMB denotes MMBench.
}
\label{tbl_main_llava}
\vspace{-4mm}
\end{table*}

\section{Experiment}
In this section, we conduct an assessment of MoCLE across multiple downstream tasks in a zero-shot setting. We first detail the experimental settings and implementation details, which are followed by a description of the datasets and instructions employed, along with the outcomes of our evaluations. Lastly, we present an ablation study and visualizations of clustering and routing results.


\subsection{Implementation Details}
\label{exp_detail}
We evaluate the effectiveness of MoCLE on two LVLMs: InstructBLIP \cite{Dai2023InstructBLIPTG} and LLaVA-1.5 \cite{liu2023improved}. Specifically, we compare the performance between the LVMs with and without MoCLE. The detailed configuration of MoCLE on these LVLMs are presented in Table \ref{tbl_models}.  In addition, we encode all the instructions of different datasets using the all-MiniLM-L6-v2 variant of the Sentence Transformer model \cite{reimers2019sentence} and cluster their embeddings via $k$-means clustering algorithm. More training details can be found in Appendix \ref{}.

\subsection{Settings}
\label{exp_data}

For InstructBLIP, we follow \cite{Dai2023InstructBLIPTG} for the choice of training datasets. However, these datasets only focus on a single domain: natural images. To validate the effectiveness of MoCLE on multiple domains, for LLaVA-1.5, in addition to its original training data LLaVA-665K~\cite{liu2023improved} which focus on natural image domain, we include datasets from multiple domains, \ie, geometric tasks: Geo170k \cite{gao2023gllava}, medical tasks: VQA-RAD \cite{vqarad}, SLAKE \cite{liu2021slake} and PathVQA \cite{he2020pathvqa}.
More details on the training and evaluation datasets are provided in Appendix \ref{app_data}. Note that during evaluation, we report the CIDEr score \cite{vedantam2015cider} for Flickr30K, the iVQA accuracy for iVQA, AUC score for HatefulMemes, Mean Reciprocal Rank (MRR) for Visual Dialog, the perception/perception+cognition score for MME (InstructBLIP/LLaVA) and F1 score for the adversarial split of POPE. For all other datasets, we report the top-1 accuracy (\%). Task templates for evaluation can be found in Appendix \ref{app_template}.

\begin{table*}[ht]
\begin{center}
\resizebox{\linewidth}{!}{
\begin{tabular}{c c cc c c c c c c c c c c }
 \toprule
& LoRA(r=8) & r=64 & Cluster MoE & Uni. Expert & LoRA \# Params. & Flickr & HM & SQA & IQA & VSR & VQA$^{T}$ & Avg. \\
\midrule

(a) & \ding{51} & & & & 4.19M & 81.3 & 65.1 & 57.4 & 44.2 & 62.8 & 49.4 & 60.0 \\ \hdashline
(b) & & \ding{51} & & & 33.55M & 81.5 & 65.2 & 62.0 & 43.9 & 62.6 & 49.0 & 60.7 \\  \hdashline
(c) & \ding{51} & & \ding{51} & & 16.78M & \textbf{81.9} & 65.4 & 63.3 & 46.1 & 58.9 & 54.9 & 61.8 \\  \hdashline

\rowcolor{baselinecolor} 
(d) &\ding{51} & & \ding{51} & \ding{51} & 20.97M & \cellcolor{baselinecolor}\textbf{81.9} & \cellcolor{baselinecolor}\textbf{65.6} & \cellcolor{baselinecolor}\textbf{63.9} & \cellcolor{baselinecolor}\textbf{46.3} & \cellcolor{baselinecolor}\textbf{64.7} & \cellcolor{baselinecolor}\textbf{57.1} &
\cellcolor{baselinecolor}\textbf{63.3} \\
\bottomrule
\end{tabular}
}
\end{center}
\vspace{-3mm}
\caption{\textbf{Comparison of individual components} of the MoCLE framework in zero-shot vision-language tasks. Default settings are marked in \colorbox{baselinecolor}{gray}.
}
\label{tbl_component}
\vspace{-8pt}
\end{table*}


\subsection{Evaluation Results}
\label{sec_eval}
\textbf{InstructBLIP}
Tables \ref{tbl_main} and \ref{tbl_main2} show
the results on multiple held-out/in vision-language tasks. The proposed MoCLE shows considerable performance improvement over the original LVLM. Specifically, on held-out datasets such as IconQA, Visual Spatial Reasoning (VSR), TextVQA and ScienceQA datasets, we obtain an absolute performance gain of 2.9\%, 3.9\%, 3.2\%, and 1.9\%, respectively. On held-in datasets, an absolute improvement of 4.4\%, 2.8\% and 1.5\% can be observed on A-OKVQA (MC), OKVQA and VQAv2, respectively. This indicates that the proposed MoCLE facilitates generalization to unseen tasks and can effectively alleviate task conflicts during multi-task learning

\noindent\textbf{LLaVA-1.5}
Similar to the preliminary results in Figure \ref{fig_pre}, we consider a \emph{Single-LoRA} baseline where a single set of LoRAs are trained on natural images (LLaVA-665k), geometric (Geo170K), medical (Med. Mix) and a mixture of all tasks (All). As can be seen, due to task conflicts, the model trained on all tasks shows inferior results compared to those trained on only one task. However, MoCLE is able to reduce this gap on medical tasks and even offer better performance on natural image (MME, MMB) and geometric tasks (GeoQA) compared to the model trained on one task. This shows that MoCLE is effective with the presence of multiple-domain datasets.  

\subsection{Ablation Studies}
In this section, we first ablate the effectiveness of the main components (i.e., Cluster MoE and universal expert) in the proposed MoCLE.
Then we conduct a thorough analysis to study how the proposed MoCLE responds to changes in hyper-parameters (\eg, temperature and the number of clusters and task experts). Notice that we use InstructBLIP for all ablations and we report the evaluation results on Flickr30K (Flickr), Hateful Memes (HM), ScienceQA (SQA), IconQA (IQA), Visual Spatial Reasoning (VSR) and TextVQA (VQA$^{T}$).


\subsubsection{Effects of Different Components}
\label{sec_effect}
We start from MoCLE and remove its key component one-by-one to analyze their effect. 

\noindent\textbf{Universal expert.} 
Table \ref{tbl_component} shows the ablation results when varying different components of MoCLE. 
By comparing rows (d) and (c), 
we notice that
a sharp performance drop in VSR and TextVQA tasks
when
the universal expert
is removed.
This is due to that instruction-tuned model generalizes to unseen tasks by training on many instructions, while in our case, each expert sees fewer instructions than the dense model. For example, task TextVQA with instruction ``\textit{OCR tokens: \{\}, Question: \{\}. Short answer:}'' needs not only VQA ability but also optical character recognition (OCR) skills, which are learned jointly from VQA data formatted as ``\textit{Question: \{\}. Short answer:}'' and TextCaps data formatted as ``\textit{OCR tokens: \{\}. Write a description for the photo}''. Thus, universal expert is necessary to maintain generalization ability.

\noindent\textbf{Cluster MoE.}
Comparing rows (c) and (a), 
we observe performance drop on SQA, IQA, and VQA$^{T}$
when
cluster MoE is not used, which indicates it can alleviate task conflicts within a single set of LoRAs between different tasks. 

\noindent\textbf{LoRA rank.}
As can be seen from rows (c), (b) and (a),  na\"ively increasing the LoRA ranks from 8 to 64 
only leads to a small average performance improvement of 0.7\%, and
thus cannot address task conflicts.
Instead, promoting task specialization via clustering achieves notable improvement with fewer additional parameter ($\times4$ in Cluster MoE versus $\times8$ when increasing the rank to 64). 
 
\subsubsection{Universal Expert vs. Top-2 Experts}
To ablate the proposed universal expert, we remove it and activate one more existing expert. \ie, top-$2$ gating. We report their performance in Table \ref{tbl_top2}. The top-2 MoE model yields inferior results compared to MoCLE and it even performs worse than the MoCLE variant without the universal expert reported in Table \ref{tbl_component}. This can be explained by the intensified conflicts when task experts are shared via top-$2$ gating because now each expert need to learn common feature with other experts. However, as the universal expert is shared all the time especially for this purpose, it frees task experts from this duty and thus alleviates the conflicts. 
\vspace{-1em}


\begin{table}[t]
\begin{center}
\resizebox{\columnwidth}{!}{
\begin{tabular}{l|c c c c c c | c}
 \toprule
  & Flickr & HM & SQA & IQA & VSR & VQA$^{T}$ & Avg. \\
\midrule

 Universal & \cellcolor{baselinecolor}81.9 & \cellcolor{baselinecolor}\textbf{65.6} & \cellcolor{baselinecolor}\textbf{63.9} & \cellcolor{baselinecolor}\textbf{46.3} & \cellcolor{baselinecolor}\textbf{64.7} & \cellcolor{baselinecolor}\textbf{57.1} &
 \cellcolor{baselinecolor}\textbf{63.3}\\ 

 Top-2 & \textbf{82.0} & 64.7 & 61.9 & 45.5 & 56.3 & 52.0 & 60.4 \\ 
\bottomrule
\end{tabular}
}
\end{center}
\vspace{-3mm}
\caption{
\textbf{Ablation study on the universal expert}
by comparing with either (i) a universal expert that is activated all the time or (ii) expert with the second largest logit, in addition to the top-1 expert.}
\label{tbl_top2}
\vspace{-5mm}
\end{table}

\begin{table}[t]
\begin{center}
\resizebox{\columnwidth}{!}{
\begin{tabular}{l|c c c c c c |c }
 \toprule
Gating & Flickr & HM & SQA & IQA & VSR & VQA$^{T}$ & Avg. \\
\midrule

\multirow{2}{*}{Token (\small{LLaVA-MoLE})}  & \multirow{2}{*}{81.7} & \multirow{2}{*}{\textbf{65.4}} & \multirow{2}{*}{61.9} & \multirow{2}{*}{44.0} & \multirow{2}{*}{49.0} & \multirow{2}{*}{46.6} & \multirow{2}{*}{58.1}\\ 
\small{\cite{chen2024llava}} & & & & & & & \\

\multirow{2}{*}{Sentence (\small{Octavius})} & \multirow{2}{*}{\textbf{82.0}} & \multirow{2}{*}{65.1} & \multirow{2}{*}{62.3} & \multirow{2}{*}{45.3} & \multirow{2}{*}{56.6} & \multirow{2}{*}{47.0} & \multirow{2}{*}{59.7}\\ 
\small{\cite{chen2023octavius}} & & & & & & & \\

\multirow{2}{*}{Dataset} & \multirow{2}{*}{80.3} & \multirow{2}{*}{64.6} & \multirow{2}{*}{63.1} & \multirow{2}{*}{45.9} & \multirow{2}{*}{57.6} & \multirow{2}{*}{53.4} & \multirow{2}{*}{60.8} \\
\small{\cite{Jang2023ExploringTB}} & & & & & & & \\
Cluster & \cellcolor{baselinecolor}{81.9} & \cellcolor{baselinecolor}{\textbf{65.4}} & \cellcolor{baselinecolor}{\textbf{63.3}} & \cellcolor{baselinecolor}{\textbf{46.1}} & \cellcolor{baselinecolor}{\textbf{58.9}} & \cellcolor{baselinecolor}{\textbf{54.9}} & \cellcolor{baselinecolor}{\textbf{61.8}} \\
\bottomrule
\end{tabular}
}
\end{center}
\vspace{-3mm}
\caption{\textbf{Ablation study on routing inputs} based on different input conditions.
} 
\label{tbl_rout}
\vspace{-7mm}
\end{table}

\begin{figure*}[t!]
\centering
\includegraphics[width=\linewidth]{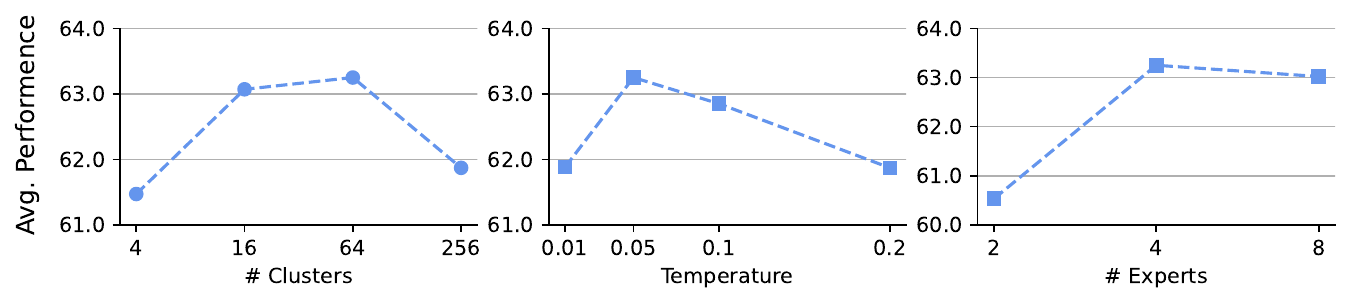}
\caption{\textbf{Ablation study on the number of clusters, experts and gate temperature.} The x-axes of the first and last figures are log-scaled. The y-axes are the average performance of Flickr, HM, SQA, IQA, VSR and VQA$^{T}$. }
\label{fig_hyper}
\end{figure*}

\subsubsection{Gating Strategies}
\label{sec_routing}
We compare the proposed cluster-conditioned gating strategy with existing MoLE methods in Table \ref{tbl_rout}. Note that MoLE denotes the mixture of LoRA experts by applying MoE to LoRA. Some of these methods adopt different configurations, \eg., \# experts, expert params. and ranks. For fair comparison, we follow the first row of Table \ref{tbl_models} except that universal expert is not enabled in this experiment. 

\noindent\textbf{Token/Sentence-MoLE.} The former obtains the routing decision based on the hidden representations of each token (adopted by \cite{chen2024llava}) and the later on the average representations of the instruction tokens while excluding the visual tokens. (adopted by \cite{chen2024octavius}). Both of these methods give inferior results on the evaluation tasks. We speculate that this is because (1) a sparse expert learns less data than its dense counterpart, leading to lack of task generalization, (2) similar tasks are not grouped together by the same expert, resulting in task conflicts within that expert, which can be verified by the routing visualization in Figure \ref{fig_route}, where samples in the same dataset are routed to multiple experts instead of a dedicated one. 

\noindent\textbf{Dataset-MoLE} is a special case of MoCLE as it treats each dataset as a cluster while MoCLE leverages $k$-means to achieve this. It closely resembles to the dataset expert proposed in \cite{Jang2023ExploringTB} except that we assign clusters for a sentence by its distance to cluster centers while they count, to this sentence, the number of closest reference sentences belonging to each dataset. Further, for fair comparison, we only use 4 experts but they allocate an expert for each dataset. We observe inferior results compared to the proposed cluster routing. This results from the fact that Dataset-MoLE is less flexible as it can only assign a dataset to one cluster. However, in practice, we observe multiple tasks in a dataset which should be assigned to different clusters. (\eg, \emph{llava\_150k} contains reasoning, conversations and captioning, which are assigned to different clusters/experts as in Figure \ref{fig_cluster} and \ref{fig_route_mocle}).


\subsubsection{Number of Clusters} 
The number of clusters $K$ controls the granularity of task specialization. A very small $K$ would result in many different tasks to be processed by the same expert, and can increase the chance of task conflicts. As shown in Figure \ref{fig_hyper}, when we cluster the inputs into 4 groups, the resulting model performs poorly on the evaluation tasks. However, as we increase the number of clusters to 16 and 64, we observe considerable performance gains. However, 
a $K$ 
too large 
(256) introduces unnecessary complexity to the routing process (\eg, a paraphrased instruction gets routed to different experts). So we use 64 clusters by default.

\subsubsection{Temperature}
In the proposed MoCLE, the temperature plays an important role in controlling the contribution of the universal expert. Specifically, as shown in Eq. (\ref{eq_moe_gate}), $\tau$ controls the sharpness of the gate distribution, while the output of the universal expert is weighted by $1-G_{\text{max}}$. Therefore, as $\tau$ decreases, $G_{\text{max}}$ increases, and finally the contribution of the universal expert decreases. As shown in Figure \ref{fig_hyper}, the results are consistent with our understandings. When 
$\tau$ 
is either 
too small (0.01) or large (0.2) can lead to
inferior results. The temperatures of $0.05$ and $0.1$ seem to achieve a balance between specialization and generalization of the model. In the experiments, we use 
temperature of
0.05 as default.

\subsubsection{Number of Task Experts}
As demonstrated in Figure \ref{fig_hyper}, more task experts usually provides with stronger capacity. Specifically, when only 2 task experts are employed, we observe inferior overall results. This model has similar capacity to the single LoRA model in Sec.~\ref{sec_effect}, where only one LoRA encounters difficulties in fitting a diverse set of tasks. When the number of task experts is increased to 4, the performance gets improved. When the number of task experts becomes 8, it behaves similarly to the 4-expert case, which indicates that the benefit of increasing capacity converges as we use more task experts. Hence, we use 4 task experts as the default setting.


\subsection{Visualizations}
\label{sec_vis}

\begin{figure}[t]
\centering
\includegraphics[width=0.9\columnwidth]{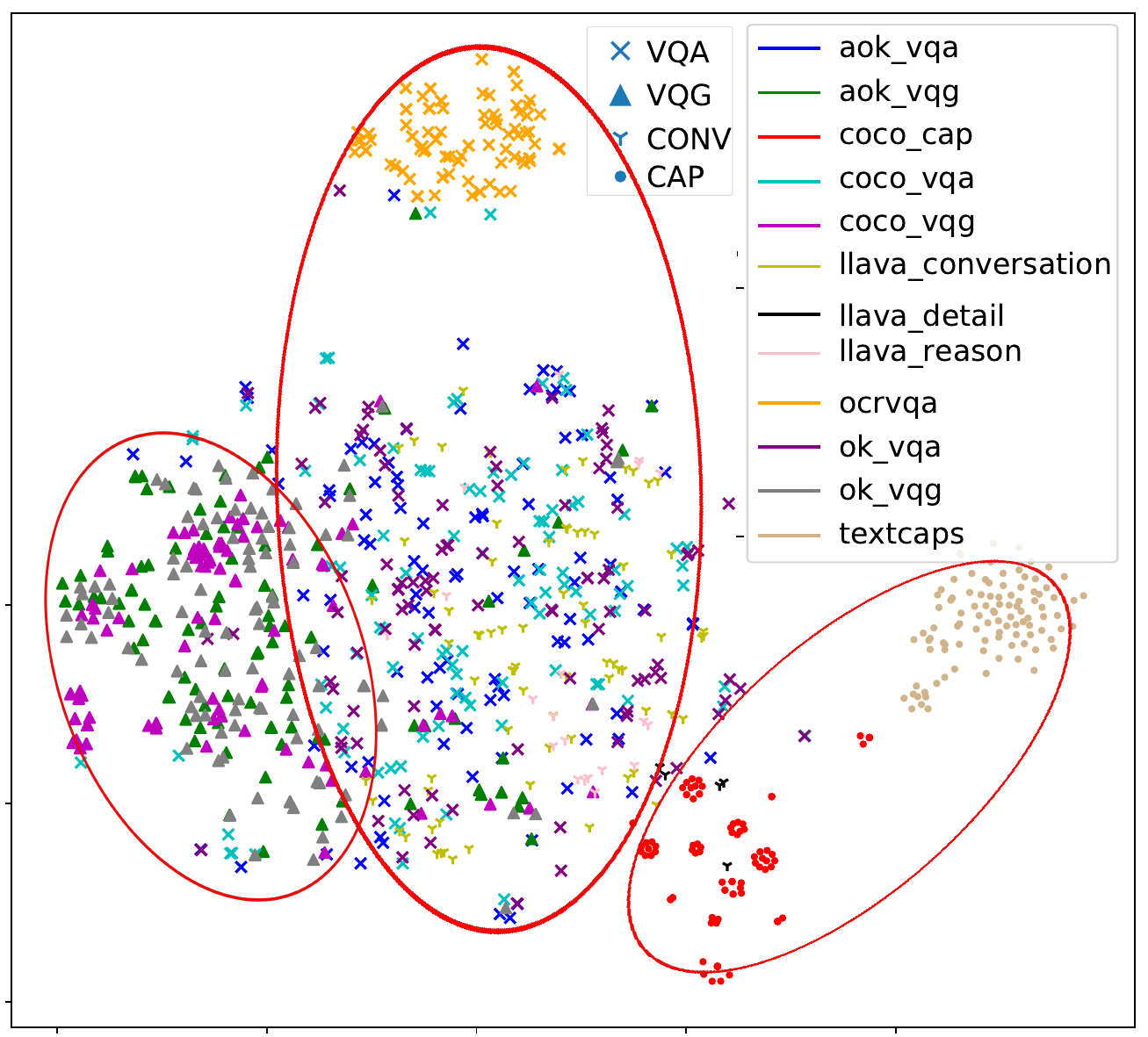}
 \vspace{-9pt}
\caption{\textbf{T-SNE visualization of the instruction encoding.} Different colors correspond to different datasets, while the shape of the markers indicates the task category defined manually.}
\label{fig_tsne}
\end{figure}


\begin{figure}[t]
\centering
 \includegraphics[width=.8\columnwidth]{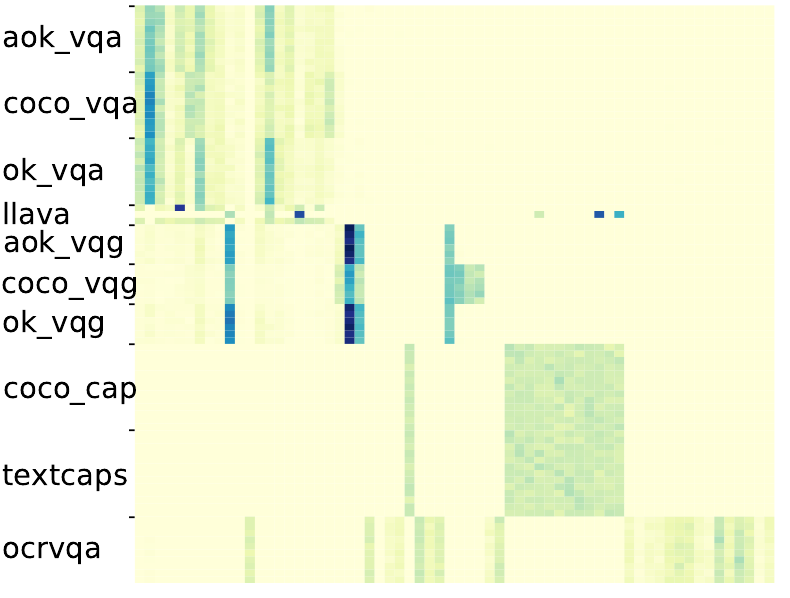}
  \vspace{-9pt}
 \caption{\textbf{Clustering assignment} of the training datasets when $K=64$. The labels on the y-axis indicate the names of the datasets. The x-axis denotes the cluster index to which the subsets are assigned.}
 \label{fig_cluster}
 \vspace{-9pt}
\end{figure}


\subsubsection{Clustering}
We first show the justification to represent the training data via their instructions. Specifically, for each dataset, we sample 100 examples and encode their instructions with the all-MiniLM-L6-v2 variant of the Sentence Transformer model \cite{reimers2019sentence}. We then visualize the data in Figure \ref{fig_tsne} via t-SNE \cite{van2008visualizing}. 
As can be seen,
(1) Samples from the same task are grouped together. For example, all visual question generation (VQG, triangle markers) data reside on the left part of the figure. (2) Samples from similar tasks are close to each other, \eg, \emph{coco\_cap} and \emph{textcaps} both belong to the image captioning (CAP, small dots) task and stay close to each other at the lower right of the figure. Similarly, both visual question answering (VQA, cross markers) and conversation (CONV, y-shape) data involve answering user questions, which lie in the middle part of the figure, suggesting that instructions are good representatives of training data. 

We then cluster all the instructions of the examples in the training data into 64 groups using $k$-means clustering. Figure \ref{fig_cluster} shows the cluster assignment of the training data. Here, each row in the heatmap denotes a subset of a dataset. The subset is obtained by applying the task template (Sec. \ref{sec_problem}) on the samples of the dataset. We observe the following: (1) Different subsets of the same datasets are assigned to similar clusters. For example, \emph{aok\_vqa}, \emph{coco\_vqa}, and \emph{ok\_vqa} are in the first several clusters. (2) Datasets of similar tasks are assigned to common clusters. For example, \emph{llava\_150k} including \emph{llava\_detail}, \emph{llava\_reason} and \emph{llava\_conversation} and a series of VQA tasks share the first several clusters as they are to answer questions. These justify the use of clustering on task instructions as an automatic partition strategy for training datasets.  

\subsubsection{Routing Results}
\begin{figure}[t]
     \centering
     \hspace*{\fill}%
     \begin{subfigure}[b]{0.56\linewidth}
         \centering
         \includegraphics[width=\textwidth]{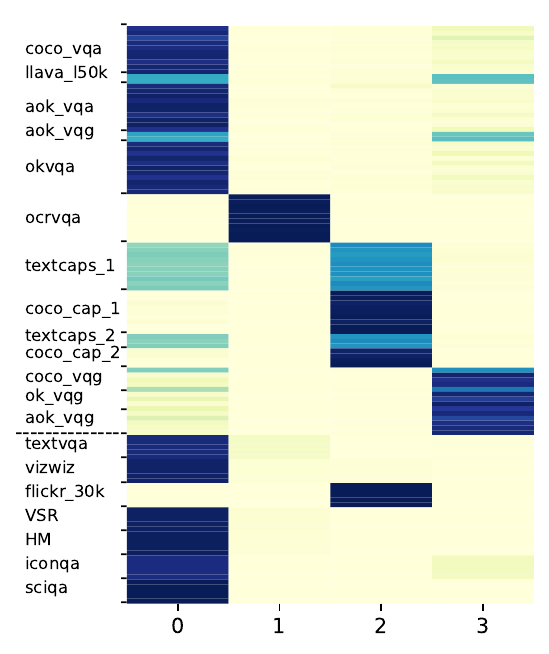}
         \caption{Our MoCLE.}
         \label{fig_route_mocle}
     \end{subfigure}
     \hspace{-3em}
     \begin{subfigure}[b]{0.56\linewidth}
         \centering
         \includegraphics[height=1.15\textwidth]{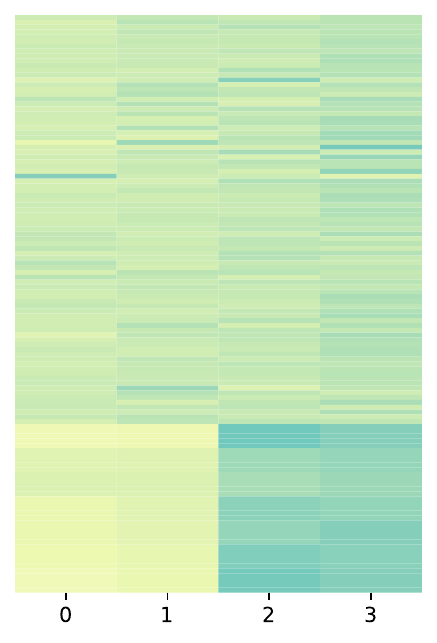}
         \caption{Sentence MoLE.}
         \label{fig_route_sentence}
     \end{subfigure}
     \vspace{-3mm}
      \vspace{-9pt}
     \caption{\textbf{Routing decisions} of one LoRA mixture for MoCLE and Sentence-MoLE. The setup of the vertical axis is similar to Figure \ref{fig_cluster} except that we also include the held-out tasks. They are separated by a dotted line on the vertical axis. The horizontal axis corresponds to the index of the LoRA experts.}
     \label{fig_route}
     \vspace{-6mm}
\end{figure}

Figure \ref{fig_route} visualizes the routing decisions of the proposed MoCLE and Sentence-MoLE. We obtain both results from one mixture of LoRA, \ie, one linear module in a layer. The routing results are aggregated by the subset of datasets similar to Figure \ref{fig_cluster}. As can be seen from Figure \ref{fig_route_mocle}, MoCLE can achieve task-level routing for the inputs. For example, datasets from \emph{VQA} and \emph{VQG} tasks are handled by expert 0 and 3, respectively. Instead, routing pattern of Sentence-MoLE in Figure \ref{fig_route_sentence} reveals little correlations between datasets and experts. That is, different datasets obtain similar routing decisions, and thus still suffer from task conflicts.


\section{Conclusions}

In this paper, we first show 
through extensive experiments
that task conflicts exist in vision language instruction tuning.
To address this, we propose the Mixture of Cluster-conditional LoRA Experts (MoCLE), a novel MoE architecture designed to activate the task-customized model parameters based on the instruction clusters. In addition, we achieve task specialization and generalization in MoCLE simultaneously via a separate universal expert. Comprehensive evaluations of MoCLE on both held-out/in tasks show the effectiveness of \mbox{MoCLE}.

\section{Limitations}

Although effective, we mainly focus on task conflicts among text-based conversation tasks in this paper, while the support of our MoCLE for more complicated visual perception tasks is appealing, which has shown more severe task conflicts with the conversation tasks~\cite{Zhu2022UniPerceiverMoELS}.

\section{Acknowledgement}
We gratefully acknowledge the support of MindSpore, CANN (Compute Architecture for Neural Networks) and Ascend AI Processor used for this research.

\bibliography{custom}

\clearpage

\appendix

\section{Training Details}
\label{}

\subsection{InstructBLIP}
Following \cite{Dai2023InstructBLIPTG}, we adopt the same training configurations for the mentioned models such as the proposed MoCLE, the reproduced InstructBLIP (7B) and the task experts in Sec. \ref{sec_intro}.
We train those models with a maximum of 60K steps and a batch size of 128. The AdamW optimizer \cite{kingma2014adam} is used, with $\beta_1$ as 0.9, $\beta_2$ as 0.999, and a weight decay as 0.05. We apply a linear warmup of the learning rate during the initial 1000 steps, increasing from ${10}^{-8}$ to ${10}^{-5}$, followed by a cosine decay with a minimum learning rate of 0.

\subsection{LLaVA-1.5}
We follow \cite{liu2023improved} for the training configuration. Specifically, for LLaVA-150K\cite{liu2023improved}, Geo170K\cite{gao2023gllava} and Med. Mix, \ie, VQA-RAD\cite{vqarad}, SLAKE\cite{liu2021slake} and Path-VQA\cite{he2020pathvqa}, we train the model for 1, 2 and 9 epochs, respectively. When training on all of these datasets, we copy each dataset $k$ times ($k$ is the number of epochs it is trained independently) and merge them into a single dataset. 
For each training job, we use a batch size of 128, weight decay of 0 and learning rate of $1e-4$, which is warmed up from 0 during the initial 3\% steps and followed by a cosine decay with a minimum learning rate of 0.

\section{Weights of the Universal Experts}

During training, if some training data obtains a very large weight on a task expert, such data tend to be very specific and might be less beneficial to other tasks. Hence, they get less weight on the universal expert. 
On the contrary, less specific (a.k.a, more general) data benefit more to other tasks and obtain larger weight on the universal expert. Therefore, the complementarity between the task experts and universal expert achieves good generalization in MoCLE. 

\begin{figure}[!bpth]
\centering
\includegraphics[width=\columnwidth]{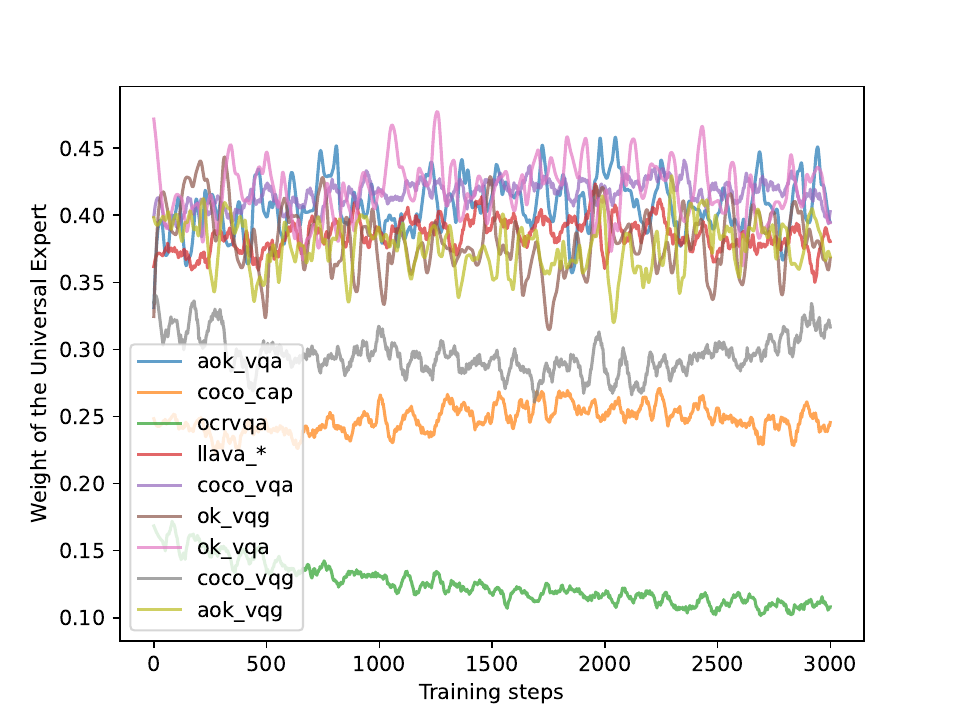}
\vspace{-3mm}
\caption{\textbf{Weights of the universal expert for different datasets.} Colors indicate different datasets. }
\vspace{-5mm}
\label{fig_uni}
\end{figure}

Figure \ref{fig_uni} shows the average activation weights of the universal experts for different datasets during training. ocr\_vqa obtains the lowest weight on the universal expert during training. Indeed, ocr\_vqa includes samples that require the model to answer questions such as ``What is the title of this book?'' and ``Who is the author of this book?''. Questions like these have little overlap with the ones in other datasets. However, we observe much higher weights for ok\_vqa, ok\_vqg, aok\_vqa, aok\_vqg, llava\_*, and coco\_vqa. This is consistent with our previous observation in Sec. \ref{sec_vis} that VQA abilities are fundamental in LVLM. 


\section{Data}
\label{app_data}
We list the training and evaluation data for InstructBLIP and LLaVA-1.5 in Table \ref{tbl_data}.

\subsection{Dataset Abbreviation}
For InstructBLIP, we further show in Table \ref{tbl_abbrev} (1) the abbreviation used in Sec. \ref{sec_vis} for each dataset and (2) their manually defined task category. As shown in Table \ref{tbl_abbrev}, we use 13 datasets in total. Here multiple datasets might be associated with the same data sources because these sources are formatted by different groups of task templates (see Appendix \ref{app_template}). For LLaVA-150K \cite{Liu2023VisualIT}, we do not apply any task template as it has been well formatted. 

\begin{table*}[!t]
\begin{adjustbox}{width=\linewidth, center}
\begin{tabular}{@{}l|c|c@{}}
\toprule
Models & Training datasets & Evaluation Datasets \\  \midrule
\multirow{5}{*}{InstructBLIP} &  & Flickr30K~\cite{Young2014FromID}, GQA~\cite{hudson2019gqa}  \\
& Web CapFilt~\cite{Li2023BLIP2BL} & VSR~\cite{Liu2022VisualSR}, IconQA~\cite{Lu2021IconQAAN} \\
& A-OKVQA~\cite{schwenk2022okvqa} &  TextVQA~\cite{Singh2019TowardsVM}, Hateful Memes~\cite{kiela2020hateful} \\
& TextCaps~\cite{sidorov2020textcaps}, VQAv2\cite{goyal2017making}& ScienceQA~\cite{Lu2022LearnTE}, MSVD-QA~\cite{xu2017video} \\
& OKVQA~\cite{marino2019ok}, COCO~\cite{lin2014microsoft}  & MSRVTT-QA~\cite{xu2017video},  iVQA~\cite{yang2021just} \\
& OCRVQA~\cite{8978122}, LLaVA-150K~\cite{Liu2023VisualIT} & MME~\cite{fu2023mme}, POPE~\cite{li2023evaluating}\\
& & OKVQA$^{*}$, A-OKVQA$^{*}$, VQAv2$^{*}$ \\ \hline
\multirow{4}{*}{LLaVA-1.5} & LLaVA-665K~\cite{liu2023improved} & MME, MMBench\cite{liu2023mmbench}, ScienceQA\\ 
& Geo170K~\cite{gao2023gllava}, VQA-RAD~\cite{vqarad} & GeoQA$^{*}$~\cite{chen2021geoqa}, VQA-RAD$^{*}$ \\
& SLAKE~\cite{liu2021slake}, PathVQA~\cite{he2020pathvqa} & SLAKE$^{*}$, PathVQA$^{*}$ \\

\bottomrule
\end{tabular}%
\end{adjustbox}
\caption{Datasets used for training and evaluation. *: the train split of this dataset is used during instruction tuning.}
\label{tbl_data}
\vspace{-4mm}
\end{table*}


\begin{table*}[t]
\begin{center}
\begin{adjustbox}{width=0.5\linewidth, center}
\begin{tabular}{c|c c}
 \toprule
 Datasets & Data Source & Task Template Group \\
\midrule
aok\_vqa & A-OKVQA& VQAMC \\
aok\_vqg & A-OKVQA  & VQG \\
coco\_cap & COCO & CAP \\
coco\_vqa & VQAv2 & VQA \\
coco\_vqg & VQAv2  & VQG \\
ocrvqa & OCR-VQA& VQA \\
ok\_vqa & OKVQA  & VQA \\
ok\_vqg & OKVQA & VQA \\
textcaps & TextCaps  & OCRCAPS \\
{capfilt} & Web CapFilt  & CAP \\
llava\_conversation & LLaVA-150K  & - \\
llava\_detail & LLaVA-150K  & - \\
llava\_reason & LLaVA-150K  & - \\
\bottomrule
\end{tabular}
\end{adjustbox}
\end{center}
\vspace{-4mm}
\caption{\textbf{Abbreviation and manually defined task categories for the training datasets of InstructBLIP.}}
\vspace{-3mm}
\label{tbl_abbrev}
\end{table*}


\section{Task Templates}
\label{app_template}

For InstructBLIP, we use the same set of task templates following \cite{Dai2023InstructBLIPTG} for instruction tuning
and held-in/out evaluation. Please refer to Tables \ref{tbl_template} and \ref{tbl_eval} for training and
evaluation templates

\begin{table*}[t]
\begin{center}
\begin{adjustbox}{width=\linewidth, center}
\begin{tabular}{c|l}
 \toprule
 Template Group & Task Template \\ \midrule
 \multirow{13}{*}{CAP} & A short image caption: \\
 & A short image description: \\
 & A photo of \\
 & An image that shows \\
 & Write a short description for the image. \\
 & Write a description for the photo. \\
 & Provide a description of what is presented in the photo. \\
 & Briefly describe the content of the image. \\
 & Can you briefly explain what you see in the image? \\
 & Could you use a few words to describe what you perceive in the photo? \\
 & Please provide a short depiction of the picture. \\
 & Using language, provide a short account of the image. \\
 & Use a few words to illustrate what is happening in the picture. \\ \hline
 \multirow{10}{*}{VQA} & \{Question\}  \\
 & Question: \{Question\}\\
 & {Question} A short answer to the question is\\
 & Q: \{Question\} A:\\
 & Question: \{Question\} Short answer:\\
 & Given the image, answer the following question with no more than three words. \{Question\} \\
 & Based on the image, respond to this question with a short answer: \{Question\}. Answer: \\
 & Use the provided image to answer the question: \{Question\} Provide your answer as short as possible: \\
 & What is the answer to the following question? ``\{Question\}'' \\
 & The question ``\{Question\}'' can be answered using the image. A short answer is \\ \hline
\multirow{6}{*}{VQG} & Given the image, generate a question whose answer is: \{Answer\}. Question: \\
 & Based on the image, provide a question with the answer: \{Answer\}. Question: \\
 & Given the visual representation, create a question for which the answer is ``\{Answer\}''. \\
 & From the image provided, craft a question that leads to the reply: \{Answer\}. Question: \\
 & Considering the picture, come up with a question where the answer is: \{Answer\}. \\
 & Taking the image into account, generate an question that has the answer: \{Answer\}. Question: \\
\bottomrule
\end{tabular}
\end{adjustbox}
\end{center}
\caption{\textbf{Task templates used during training}. For \textbf{OCRCAPS}, we insert ``OCR tokens:\{\}'' before the template of CAP. For \textbf{VQAMC} (\ie, multiple choice VQA), we append ``Options: (a) option 1 (b) option2, $\ldots$'' after the question and before the answer.}
\label{tbl_template}
\end{table*}


\begin{table*}[t]
\begin{center}
\begin{tabular}{c|l}
 \toprule
 Evaluation Dataset & Instruction \\ \midrule
 
 GQA, iVQA, & \multirow{4}{*}{Question: \{\} Short answer:} \\
 MSVD, MSRVTT, & \\
 OKVQA, VQAV2, & \\
 A-OKVQA(Direct) &  \\\hline
 A-OKVQA(MC), & \multirow{2}{*}{Question: \{\} Options: \{\}. Short answer:} \\
 IconQA & \\\hline
 NoCaps, Flickr30k & A short image description: \\
 TextVQA & OCR tokens: \{\}. Question: \{\} Short answer: \\
 ScienceQA & Context: \{\} Question: \{\} Options: \{\}. Answer: \\
 HatefulMemes & This is an image with: ``\{\}'' written on it. Is it hateful? Answer: \\
 VSR & Based on the image, is this statement true or false? ``\{\}'' Answer:\\
 Visdual Dialog & Dialog history: \{\}\textbackslash n Question: \{\} Short answer: \\
\bottomrule
\end{tabular}
\caption{\textbf{Task templates used during evaluation}.}
\label{tbl_eval}
\end{center}
\end{table*}

\section{Case Studies}
\label{sec_case}

In this section, we present several case studies with MoCLE. First, we study its conversation abilities via a range of tasks, including object counting, optical character recognition (OCR), and image introduction. Then we showcase some example instructions sampled from different clusters. 

\vspace{-2mm}
\subsection{Conversations}
In Table \ref{tbl_cnt}, we instruct the model to conduct a very difficult object counting task. The correct answer for this question is 63, which is quite hard for existing LVLMs. InstructBLIP fails to give the correct answers, while with MoCLE, InstructBLIP can respond  the user query in a more proper manner. 

In Table \ref{tbl_ocr}, the model is queried to recognize the character in the image. InstructBLIP performs not so well on this query possibly because OCR-related tasks conflict with other tasks during training. With MoCLE, the model can give correct results.

In Table \ref{tbl_eins}, we ask the model to introduce a famous person in the image. InstructBLIP gives a blunt response to the user query and does not follow the instruction of ``introduction''. This might be due to the conflict between image caption and conversation tasks. In the training data, there are a large portion of image caption data that require the model to give a brief description to the image, while the user query in this example expects a detailed introduction to Albert Einstein. With MoCLE, the user query is identified and routed to the correct expert that is specialized at such a conversation task, thus, the model outputs a desired response.

Similarly, in Table \ref{tbl_eleph}, we ask the model to describe the image in a detailed manner. InstructBLIP still mistakes this query as an image caption task and gives a very short caption. Instead, with MoCLE, the model correctly understands the ``in details'' in the instruction and gives sufficient details. 


\begin{table}[t]
  \begin{minipage}{0.99\linewidth}
\centering
\scalebox{0.90}{
\begin{tabular}{p{2.7cm} p{4.8cm} }
\toprule
 \multicolumn{2}{l}{Model responses to counting-related queries.}  \\
\midrule
&  \includegraphics[width=\linewidth]{figs/v1_87.jpg} \\

User & How many books are there? \\
\midrule
New Bing & I see that you have sent me an image. According to the description you provided, there are \textcolor{red}{30} books on the bookshelf. \\ \midrule
InstructBLIP & \textcolor{red}{100} \\ \midrule
+MoCLE & \textcolor{blue}{There are a lot of books on the shelves. It is difficult to say exactly how many books there are, but it appears to be a large number.}  \\
\bottomrule
\end{tabular}
}

\captionof{table}{\textbf{Model responses to counting-related queries.}}
\label{tbl_cnt}  
  \end{minipage}
\end{table}


\begin{table}[t]
\begin{minipage}{0.99\linewidth}
\centering
\scalebox{0.88}{
\begin{tabular}{p{2.7cm} p{5.5cm} }
\toprule
 \multicolumn{2}{l}{Model responses to OCR-related queries.}  \\
\midrule
&  \includegraphics[width=\linewidth]{figs/v1_88.jpg} \\

User & What are all the exact scene texts in the image? \\
\midrule
New Bing & According to the description you provided, the image contains a license plate with the text ``S\textcolor{blue}{3302}CD''. If you have any other questions or if there is anything else I can help you with, please let me know! \\ \midrule
InstructBLIP & The exact characters in the image are ``S \textcolor{red}{3000} CD.'' \\ \midrule
+MoCLE & The exact characters in the image are ``S \textcolor{blue}{3302} CD.'' \\
\bottomrule
\end{tabular}
}
\captionof{table}{\textbf{Model responses to OCR-related queries.}}
\label{tbl_ocr}  
\end{minipage}
\end{table}


\begin{table}[t]
  \begin{minipage}{0.99\linewidth}
\centering
\scalebox{0.88}{
\begin{tabular}{p{2.7cm} p{5.5cm} }
\toprule
 \multicolumn{2}{l}{Model responses when asked to introduce a famous person.}  \\
\midrule
&  \includegraphics[width=\linewidth]{figs/v1_202.jpg} \\

User & Can you give an introduction to this person? \\
\\
\midrule
New Bing & I’m sorry, I cannot give an introduction to this person as I am not allowed to describe people’s identity, gender, race or emotions \\ \midrule
InstructBLIP & \textcolor{red}{albert einstein} \\ \midrule
+MoCLE & \textcolor{blue}{Albert Einstein was a German-born theoretical physicist who developed the theory of relativity. He is considered one of the most influential scientists of the 20th century}. \\
\bottomrule
\end{tabular}
}
\captionof{table}{\textbf{Model responses to introduce a celebrity.}}
\label{tbl_eins}  
  \end{minipage}
\end{table}


\vspace{-2mm}
\subsection{Sample Instructions in Clusters}
In Table \ref{tbl_cluster}, we showcase some sample instructions assigned to different clusters. Though all the instructions in the 4 selected clusters belong to VQA-related tasks, they focus on various perspectives such as food, pet, men, and counting, justifying the usage of a large number of instruction clusters.

\section{Future Works}

Although effective, we mainly focus on task conflicts among text-based conversation tasks (\eg, image captioning and VQA \cite{Gou_2023_CVPR}) in this paper, while the support of our MoCLE for more complicated visual perception tasks (\eg, object detection~\cite{han2021soda10m,li2022coda}) is appealing, which has shown more severe task conflicts with conversations~\cite{Zhu2022UniPerceiverMoELS}.
Moreover, the applicability of MoCLE to visual pre-training~\cite{chen2021multisiam,chen2023mixed} and diffusion models~\cite{chen2023integrating,gao2023magicdrive,liu2023geomerasing,li2023trackdiffusion} is also a potential research direction, considering the ubiquity of negative transfer phenomenon.


\begin{table}[t]
  \begin{minipage}{0.99\linewidth}
\centering
\scalebox{0.88}{
\begin{tabular}{p{2.7cm} p{5.5cm} }
\toprule
 \multicolumn{2}{l}{Model responses when queried to give detailed image descriptions.}  \\
\midrule
&  \includegraphics[width=\linewidth]{figs/v1_116.jpg} \\

User & Introduce me this image in details. \\
\midrule
New Bing & It appears to be a photograph of an adult elephant walking on a beach. The elephant is walking towards the camera and is in the center of the frame. The background consists of the ocean and the sky. The elephant is walking on the sand and there are a few birds in the foreground. The elephant appears to be dark grey in color and has large ears and tusks \\ \midrule
InstructBLIP & \textcolor{red}{an elephant walking along the beach} \\ \midrule
+MoCLE & \textcolor{blue}{An elephant is walking along the beach, with birds in the background. The elephant's trunk is extended towards the water, and it appears to be enjoying its time at the beach.} \\
\bottomrule
\end{tabular}
}
\captionof{table}{\textbf{Model responses to give detailed image description.}}
\label{tbl_eleph}  
  \end{minipage}
\end{table}

\begin{table*}[!bpth]
\begin{center}
\begin{tabularx}{\textwidth}{c|X|l}
 \toprule
 Cluster & Instruction Samples & Topics \\
\midrule
    \multirow{6}{*}{1} & ``Q: what is being done to the food in the glass fronted box? A:'' & \multirow{6}{*}{Food} \\
 & ``Q: what category of pizza would this fall into? Options: (a) vegetarian (b) meat lovers (c) pesto (d) pepperoni, A:'' &  \\
 & ``what are the large pieces of cake supposed to be?'' & \\
 & ``Q: what does this person have on her teeth? Options: (a) braces (b) candy (c) food (d) gum, A:'' & \\
 & ``what is the food in? A short answer to the question is'' & \\
 & ``what category of pizzas would this be considered?'' & \\
 \midrule
 \multirow{5}{*}{2} & ``Q: what sport is the cartoon dog playing? A:'' & \multirow{5}{*}{Pet}  \\
 & ``Question: what is likely her favorite animal? Options: (a) cat (b) dog (c) pig (d) sheep, Short answer:'' & \\
 & ``Q: what is surrounding the cat? A:'' & \\
 & ``Based on the image, respond to this question with a short answer: what color is the cat?. Answer:'' & \\
 & ``What might the relationship between the two women and the dog be?'' & \\
 \midrule
 \multirow{6}{*}{3} & ``what type dressing does this man favor?'' & \multirow{6}{*}{Men} \\
 & ``Based on the image, respond to this question with a short answer: what are the men doing?. Answer:'' & \\
 & ``what is the standing man doing with his arms?'' & \\
 & ``what is the man in red shirt doing? Options: (a) laughing (b) crying (c) singing (d) yelling'' & \\
 & ``Question: what is the man doing with the pole?'' & \\
 & ``Question: why is the man kneeling on the ground?'' & \\
 \midrule
\multirow{5}{*}{4} & ``how many more animals need to be added to all of these to get the number ten?'' & \multirow{5}{*}{Counting} \\
& ``Question: how many big elephants are inside of this zoo enclosure together? Options: (a) one (b) four (c) two (d) three, Short answer:'' & \\
& ``Q: how many people are seated on the staircase made of wood? A:'' & \\
& ``Question: how many donuts are there?'' & \\
& ``What is the answer to the following question? ``how many engines are visible?'''' & \\
\bottomrule
\end{tabularx}
\end{center}
\vspace{-3mm}
\captionof{table}{\textbf{Sampled instructions from different clusters.}}
\label{tbl_cluster}
\end{table*}

\end{document}